%% file: CameraReady.tex
\documentclass[10pt,twocolumn,letterpaper]{article}

\usepackage{cvpr}              

\makeatletter
\@namedef{ver@everyshi.sty}{}
\makeatother
\usepackage{tikz}
\usepackage{comment}
\usepackage{amsmath,amssymb} 
\usepackage{color}

\usepackage{graphicx}

\usepackage{bm}
\usepackage{multirow}
\usepackage{multicol}
\usepackage{makecell}

\usepackage[english]{babel}

\usepackage{algorithm} 
\usepackage{algpseudocode} 
\usepackage{booktabs}
\usepackage{caption}
\usepackage{subcaption}
\usepackage{rotating}
\usepackage{verbatim}
\usepackage{gensymb}
\usepackage{wrapfig}

\usepackage{pgffor}

\usepackage{xspace}
\usepackage{enumitem}

\usepackage[pagebackref,breaklinks,colorlinks]{hyperref}

\usepackage[capitalize]{cleveref}
\crefname{section}{Sec.}{Secs.}
\Crefname{section}{Section}{Sections}
\Crefname{table}{Table}{Tables}
\crefname{table}{Tab.}{Tabs.}

\usepackage[font=small,skip=3pt]{caption}
\def\cvprPaperID{5236} 
\def\confName{CVPR}
\def\confYear{2023}

\begin{document}

\title{Tensor4D : Efficient Neural 4D Decomposition for High-fidelity Dynamic Reconstruction and Rendering}

\author{Ruizhi Shao$^{1}$, Zerong Zheng$^{1,2}$,  Hanzhang Tu$^{1}$,  Boning Liu$^{1}$,  Hongwen Zhang$^{1}$, Yebin Liu$^{1}$\\
$^{1}$Department of Automation, Tsinghua University $^{2}$NNKosmos Technology 
}

\twocolumn[{%
\renewcommand\twocolumn[1][]{#1}%
\maketitle
\begin{center}
    \centering
    \captionsetup{type=figure}
    \includegraphics[width=1\textwidth]{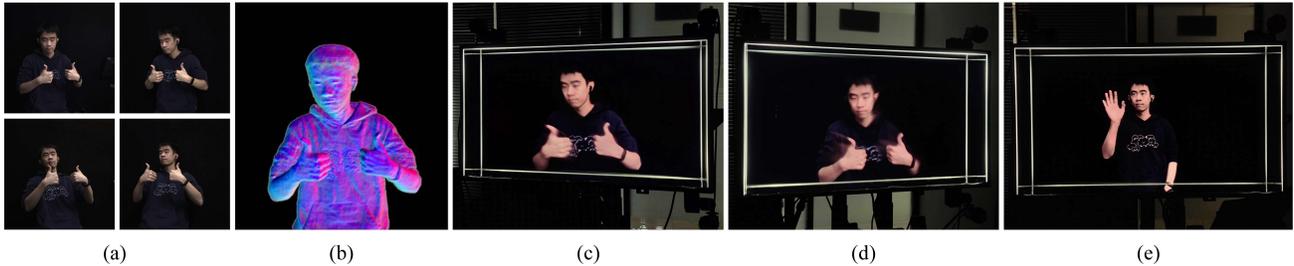}
    \vspace{-5mm}
    \captionof{figure}{Given 4 sparse static RGB camera views of a dynamic scene (a), our proposed Tensor4D decomposition enables multiview reconstruction to achieve fine-grained geometry reconstruction even on human fingers (b) and temporal-consistent novel view synthesis on a 3D holographic display (c,d,e). The 4 cameras are settled on four conners of the display. The proposed method demonstrates low-cost, portable and highly immersive telepresence experience.}
    \label{figs:teaser}
\end{center}%
}]

\begin{abstract}

\input{tex/abstract}
\end{abstract}


\input{tex/introduction}

\input{tex/relatedwork}

\input{tex/method}

\input{tex/experiment}

\input{tex/discussion}

{\small
\bibliographystyle{ieee_fullname}
\bibliography{egbib}
}

\end{document}

%% file: tex/abstract.tex
We present Tensor4D, an efficient yet effective approach to dynamic scene modeling. 
The key of our solution is an efficient 4D tensor decomposition method so that the dynamic scene can be directly represented as a 4D spatio-temporal tensor.
To tackle the accompanying memory issue, we decompose the 4D tensor hierarchically by projecting it first into three time-aware volumes and then nine compact feature planes.
In this way, spatial information over time can be simultaneously captured in a compact and memory-efficient manner.
When applying Tensor4D for dynamic scene reconstruction and rendering, we further factorize the 4D fields to different scales in the sense that structural motions and dynamic detailed changes can be learned from coarse to fine.
The effectiveness of our method is validated on both synthetic and real-world scenes.
Extensive experiments show that our method is able to achieve high-quality dynamic reconstruction and rendering from sparse-view camera rigs or even a monocular camera. The code and dataset will be released at \url{https://github.com/DSaurus/Tensor4D}.

%% file: tex/introduction.tex
\section{Introduction}
High quality reconstruction and Photo-realistic rendering of a dynamic scene from a set of input images is necessary for many applications such as AR/VR, 3D content production and entertainment. 
Traditional methods use classical mesh-based representation to reconstruct the dynamic scenes, which, unfortunately, are prone to produce reconstruction errors and rendering artifacts when the scene contains thin structures, specular surfaces and topological changes~\cite{Li2012Temporal,Virtualizedreality1997,TheRelightables2019,Fvv2015,Starck2007Surface}. 

Recent advances in neural rendering approaches, which learn scene representations in the form of neural radiance fields (NeRF), have shown impressive novel view synthesis of general static scenes given only multi-view images~\cite{nerf2020}. They are immediately extended to dynamic scenes: some methods (\textit{e.g.}, NeRF-T) consider time as an
additional input dimension to NeRF representation~\cite{VideoNeRF2021-nfds77,DCT-NeRF2021-nfds71}, while other methods (\textit{e.g.}, D-NeRF) disentangle a dynamic scene into a canonical radiance field and a dynamic motion field~\cite{D-Nerf2021-nfds54,NR-Nerf2021-nfds67,NeRFlow2021-nfds12,Nerfies2021-nfds49,DeVRF2022-nfds32}. 
Either way, learning a 4D function is one of the main cornerstones. Unfortunately, directly using MLP to fit such a function often suffers from high time and computation cost, \textit{i.e.}, dozens of hours on high-end GPUs. 

In fact, the aforementioned limitation also exists in conventional NeRF-based methods for static scenes, and researchers have proposed to use discrete data structures like voxel grids~\cite{Plenoctrees2021-nfds81} or triplanes~\cite{eg3d2022} to accelerate NeRF training and rendering. However, these techniques are difficult to be extended to dynamic domains as introducing an additional time dimension will exponentially increase memory footprint, hindering them from modeling high-quality appearance details.

In this work, we pursue a dynamic scene representation that also utilizes explicit feature grids to accelerate network training while avoiding huge memory consumption when introducing an additional time dimension. To this end, we bypass the construction of a high resolution 4D tensor; instead, we propose to model a 4D field using hierarchical tri-projection decomposition. Our decomposition method extends the tri-projection in EG3D~\cite{eg3d2022}. It firstly project a full 4D field into three time-aware volumes, each of which is then further decomposed into three feature planes. In this way, we model the 4D field using only nine 2D feature planes, and we empirically find that although being highly compact, such a representation is powerful enough to represent dynamic scenes containing complex motions. Moreover, the usage of explicit data structure also allows us to design a coarse-to-fine strategy to further improve the performance of our method. 

By utilizing and factorizing an explicit 4D tensor, our method enables both efficient reconstruction and compact representation of dynamic scenes. Besides, the decomposition scheme also introduces implicit constraints on the representation since only low-rank tensors can be approximated by a small number of lower-dimensional components. Such a constraint can serve as an inherent regularization when the input observation is limited, \textit{e.g.}, under sparse and fixed cameras setting or even monocular inputs. 
In this paper, we first apply our method for sparse-view dynamic reconstruction by adopting our Tensor4D decomposition to time-conditioned radiance fields in ``NeRF-T''.  
In addition, our decomposition method can also be used for single-view dynamic reconstruction. This is achieved through decomposing both the 4D dynamic motion field and the canonical radiance field in ``D-NeRF''.  
With proper regularization, our system enables efficient and high-quality reconstruction of dynamic objects under both camera settings.

%% file: tex/relatedwork.tex
\section{Related Work}
\noindent \textbf{Multiview Reconstruction and Rendering. }
There are various ways to capture and reconstruct the 3D dynamic scenes, involving methods based on silhouette \cite{Outdoor2012-19,freely2011-53}, stereo \cite{LRDS2018-31,shao2022diffustereo}, flow\cite{gotardo2015photogeometric, zhou2016view, shao2022floren}, segmentation \cite{DMDE2016-44,Videopop-up2014-46}, and photometric \cite{Robustfusion2008-1,DSC2009-56}. 
With RGBD cameras, real-time solutions like DynamicFusion \cite{DynamicFusion2015-td3r35} estimates the non-rigid deformations of a dynamic scene and integrates depth frames to reconstruct the geometry model in the canonical space. This method is later extended for telepresence and holographic communication~\cite{Motion2fusion2017-td3r7,ProjectStarline2021,yu2021function4d}. However, these systems heavily rely on depth sensors to obtain accurate geometry. In contrast, our method can reconstruct and render dynamic scenes using sparse-view RGB cameras.

In the past few years, neural implicit representations underwent rapid development and have been applied for multi-view reconstruction and rendering of static scenes. 
Some methods represent the geometry as the zero level-set of a neural network, and use differentiable surface rendering to optimize the network weights~\cite{DVR2020CVPR,srns2019,idr2020}. Given dense observation of an object, these methods are able to accurately recover its surface. NeRF~\cite{nerf2020}, on the other hand, uses volume rendering to optimize the scene representation. Its simplicity and impressive results inspires a lot of following works, including in-the-wild reconstruction~\cite{martinbrualla2020nerfw,sun2022neuconw}, lighting and material estimation~\cite{boss2021nerd,nerv2021,NeRFactor2021}, generation~\cite{gu2022stylenerf,jang2021codenerf,chanmonteiro2020pi-GAN,GRAF2020,sun2022fenerf,sun2022ide}, human rendering~\cite{shao2022doublefield,Zhao_2022_HumanNeRF,sun2021HOI-FVV,IButter2021,zheng2022structured} and so on. More recently, several methods unify surface and volumetric rendering, enabling accurate geometry reconstruction and high-quality novel view synthesis~\cite{volsdf2021,Unisurf2021ICCV,neus2021}.
Compared to these static scene representations, we aim to enable free view synthesis of dynamic scenes from extremely sparse cameras.

\noindent \textbf{NeRF for Dynamic Scenes.} 
Modeling scenes in 4D domain with time dimension included is a direct solution to extend NeRF for dynamic domains. Typical approaches includes VideoNeRF \cite{VideoNeRF2021-nfds77}, NeRFlow \cite{NeRFlow2021-nfds12}, DyNeRF \cite{DyNeRF2022-nfds30}, and DCT-NeRF \cite{DCT-NeRF2021-nfds71}. 
Specifically, VideoNeRF \cite{VideoNeRF2021-nfds77} directly learns a spatiotemporal irradiance field from a single video and uses depth estimation to address the shape-motion ambiguities in monocular inputs, while NeRFlow \cite{NeRFlow2021-nfds12} and DCT-NeRF \cite{DCT-NeRF2021-nfds71} use point trajectory to regularize the network optimization. 
To deal with the limitation of topology-change modeling in deformation field, Park \textit{et al.} \cite{HyperNeRF2021-nfds50} represent HyperNeRF which can lift NeRF to higher dimensions. 

Dynamic scenes can also be rendered by deforming the radiance field in the canonical space. For example, Nerfies \cite{Nerfies2021-nfds49} optimizes an additional continuous deformation field by warping each observed point into a canonical 5D NeRF. D-Nerf \cite{D-Nerf2021-nfds54} and NR-Nerf \cite{NR-Nerf2021-nfds67} follow a similar framework, but take only monocular videos as training data. 
In addition, DeVRF \cite{DeVRF2022-nfds32} uses voxel-based representation instead of MLPs to model the 3D canonical space and the 4D deformation field. 
Using parametric body templates as the semantic prior, methods like Neural Body~\cite{peng2021neuralbody} and HumanNeRF~\cite{weng_humannerf_2022_cvpr} enable photo-realistic novel view synthesis of complex human performance. 
The discrepancy between the practical capture process and the existing experimental protocols for monocular videos has been shown in \cite{MDVS2022-nfds15}. 

\noindent \textbf{NeRF Acceleration.} Numerous works emerge with the purpose of speeding up static NeRF using explicit data structures including feature maps, voxels and tensors. 
For instance, DVGO \cite{DVGO-na63} achieves fast convergence through an explicit representation of a density voxel grid and a feature voxel grid. With the sparse voxel octree structure, NSVF \cite{NSVF2020-na33} accelerates the novel view reconstruction by discarding the empty voxels in a coarse-to-fine manner.  
Similarly, Plenoxels \cite{Plenoxels2022-na57} and PlenOctree \cite{Plenoctrees2021-nfds81} model a scene through a hierarchical 3D grid with spherical harmonics, which can realize an optimization with two orders of magnitude faster than NeRF. 
In DIVeR\cite{DIVeR2022-na75}, ray marching only finds a fixed number of hits on the voxel grid to accelerate volumetric rendering. 
Moreover, hashing encoding \cite{InstantNGP2022-na44} and tensor decomposition \cite{TensoRF2022-na9} are also used as compact representations for NeRF acceleration. 

For dynamic scene modeling, 
DeVRF \cite{DeVRF2022-nfds32} enables fast non-rigid neural rendering with both 3D volumetric and 4D voxel field. In addition, V4D \cite{V4D2022-na14} introduce an effective conditional positional encoding for 4D data to realize fast novel view synthesis. 
TiNeuVox \cite{TiNeuVox2022-na13} represents scenes with optimizable explicit data structures and accelerates radiance fields modeling, while Wang \textit{et al.} extended PlenOctrees \cite{Plenoctrees2021-nfds81} into free-view video rendering \cite{Fourierplenoctrees2022-nfds72}. 
However, the low-res volumetric design of these works hinders the capacity of rendering high-quality images. In addition, there are also several concurrent works~\cite{fridovich2023kplanes, cao2023hexplane,jang2022dTensoRF} adopt 6-plane decomposition for dynamic scenes. Compared with these methods, our hierarchical decomposition is more efficient to capture time variations and represent dynamic scenes.

%% file: tex/method.tex
\section{Method}

\begin{figure}
    \centering
    \includegraphics[width=1.0\linewidth]{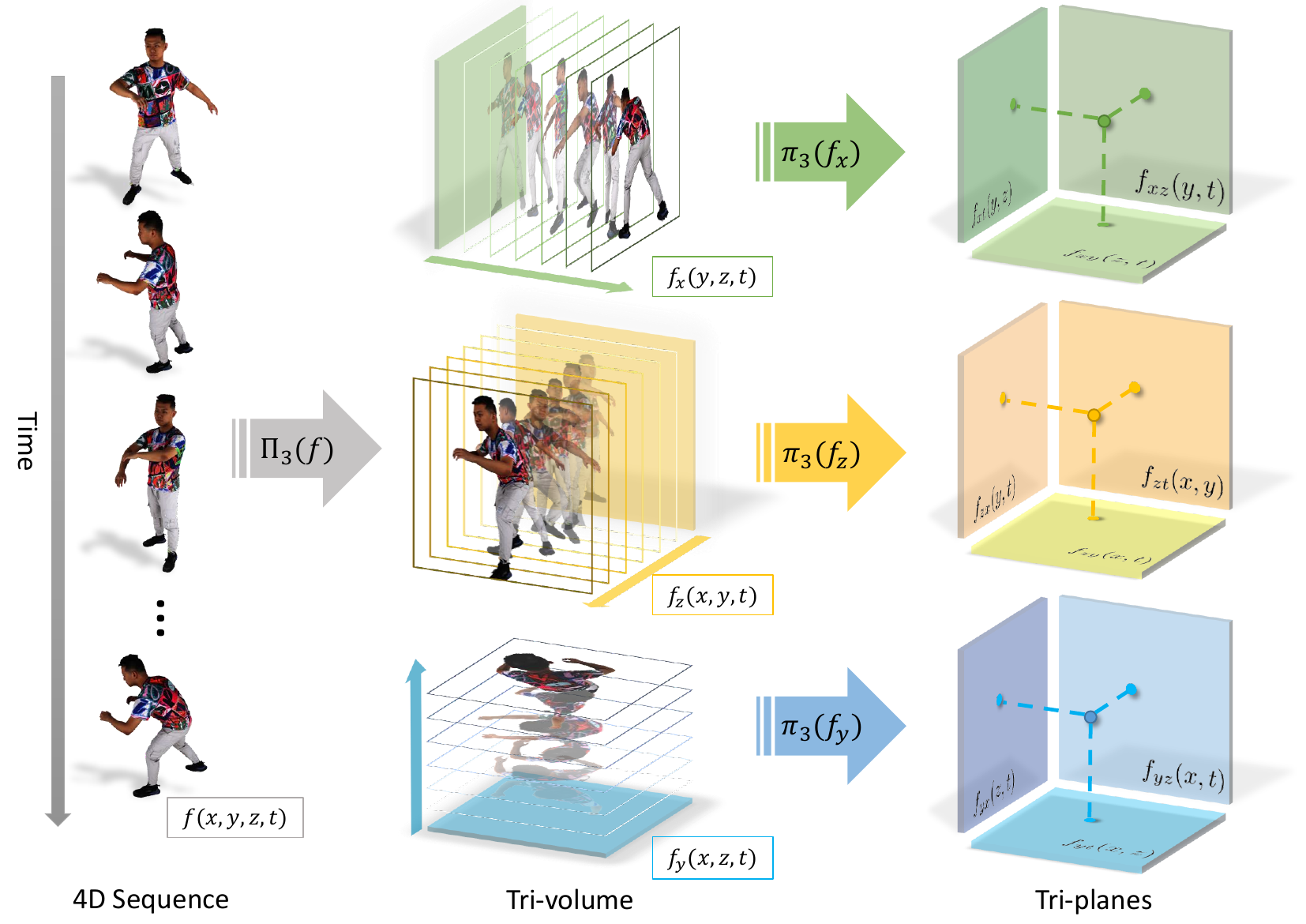}
    \caption{\textbf{Illustration of our hierarchical tri-projection decomposition method.} For a neural 4D field $f(x, y, z, t)$, we first decompose the 3D space part from 4D spatio-temporal tensor into three time-aware volumes, which are then further projected onto nine 2D planes. }
    \label{fig:reprez}
    \vspace{-3mm}
\end{figure}

Building on prior work for spatio-temporal representations and tri-projection decomposition (Sec.~\ref{sec:4d_representation}), we propose a hierarchical tri-projection decomposition method (Sec.~\ref{sec:4d_decomposition}) and a coarse-to-fine  strategy (Sec.~\ref{sec:coarse_to_fine}), which allow us to learn a 4D field at a modest cost of training time and GPU memory.

\subsection{Preliminary}
\label{sec:4d_representation}
\noindent\textbf{Spatio-temporal 4D NeRF representation.} To represent dynamic objects using neural radiance fields, a naive way is to condition the original neural radiance fields on the timestamp~\cite{VideoNeRF2021-nfds77,DCT-NeRF2021-nfds71}, which we term as NeRF-T. Mathematically, NeRF-T can be formulated as: 
\begin{equation}
\begin{split}
    f(x, y, z, t) &= (\mathbf{f}, \sigma), \\
    g(\mathbf{f}, \theta, \phi) &= \mathbf{c}, 
\end{split}
\end{equation}
where $f$ is a dynamic implicit field that produces a high-dimensional feature $\mathbf{f}$ and a density value $\sigma$ for a point at position $(x, y, z)$ and time instance $t$, and $g$ is a function that takes into account the viewing direction $(\theta, \phi)$ to predict the final RGB color. 

To achieve better disentanglement of shape and motion, some methods like D-NeRF~\cite{D-Nerf2021-nfds54} propose deformable neural radiance field which adopts a canonical 3D representation with the 4D flow fields: 
\begin{equation}
\begin{split}
    f(x, y, z, t) &= (\hat{x}, \hat{y}, \hat{z}), \\
    g(\hat{x}, \hat{y}, \hat{z}, \theta, \phi) &= ( \mathbf{c}, \sigma),
\end{split}
\end{equation}
where $g$ is the radiance field in canonical configuration and $f$ is a scene flow field representing the mapping between the scene at time instant $t$ and the canonical space.

From the above formulation one can easily observe that both NeRF-T and D-NeRF rely on modeling a 4D field $f(x, y, z, t)$, \textit{i.e.}, the dynamic implicit field in NeRF-T and the flow field  in D-NeRF. Existing methods mainly adopt MLPs to fit these 4D fields. Such an implicit neural representation does not have an explicit structure and requires extensive computation time for both training and rendering.

\noindent \textbf{Tri-projection Decomposition.} The tri-projection decomposition is widely used in recent work including EG3D~\cite{eg3d2022} and TensoRF~\cite{TensoRF2022-na9} in order to accelerate the training and rendering process in MLP-based NeRF frameworks. Such decomposition factorizes a $n$-dimensional tensor $V_h$ into three lower-dimensional ($(n-1)$-D) tensors $V_l^i\{i=1,2,3\}$ by projecting $V_h$  along the $x$, $y$ and $z$-axis respectively. For example, triplane-based decomposition proposed by EG3D projects the 3D tensor into three 2D feature planes.
Compared to voxel-based representations, triplane representation effectively reduces the memory footprint and improves the performance of 3D generation and reconstruction.

\subsection{Hierarchical Tri-projection Decomposition}
\label{sec:4d_decomposition}

Our goal is to design a dynamic scene representation with explicit feature grids to accelerate network training and volumetric rendering. However, directly constructing a 4D tensor costs a huge amount of memory and is unacceptable for the purpose of high-resolution rendering. Therefore, we propose a hierarchical triprojection decomposition to factorize the 4D tensor into several compact features, which reduces memory consumption by a large margin while preserving the capability of fitting 4D fields.  

Specifically, for a neural 4D field $f(x, y, z, t)$, 
we first decompose the 3D space part from 4D spatio-temporal tensor into three time-aware volumetric tensors via tri-projection decomposition:
\begin{equation}
\label{equ:triproj_decomp_4d}
    \Pi_3 (f(x, y, z, t)) = \left\{ f_{z}(x, y, t), f_{y}(x, z, t), f_{x}(y, z, t) \right\},
\end{equation}
where $\Pi_3$ denotes the projection operator. 
To further lower space complexity and enable high-resolution representation, 
we decompose each feature volume into three feature planes as: 
\begin{equation}
\label{equ:triproj_decomp_3d}
\begin{split}
    \pi_3 (f_{z}) &= \left\{ f_{zt}(x, y), f_{zy}(x, t), f_{zx}(y, t) \right\} \\
    \pi_3 (f_{y}) &= \left\{ f_{yt}(x, z), f_{yz}(x, t), f_{yx}(z, t) \right\} \\
    \pi_3 (f_{x}) &= \left\{ f_{xt}(y, z), f_{xz}(y, t), f_{xy}(z, t) \right\} \\
\end{split}
\end{equation}
where $\pi_3$ denotes the volume-to-plane tri-projection. In this way, we compactly represent a 4D field using 9 planes. Given any spatio-temporal coordinate $(x, y, z, t)$, we can efficiently query its value in the 4D field by projecting it onto the planes and retrieving the corresponding value via bilinear interpolation. Fig~\ref{fig:reprez} is an illustration of our decomposition method. Our hierarchical triprojection decomposition reduces the space complexity from $\mathcal{O}(n^4)$ to $\mathcal{O}(n^2)$ with $n$ being the spatial resolution of the grid, significantly lowering memory footprint without sacrificing representation power. 

\noindent \textbf{Differences from 6-plane decomposition.} Compared with 6-plane decomposition~\cite{fridovich2023kplanes, cao2023hexplane,jang2022dTensoRF} which pairs the time dimension with only one spatial dimension ($xt, yt, zt$), our method first decomposes the 4D tensor in the spatial domain to obtain three time-aware volumes and then 9 decomposed planes. Note that the three volumes are decomposed independently, making the 9 planes different from each other. Specifically, when using a linear layer $g=\{\}$ to decompose the volume $f_z(x, y, t) = \{., f_{zy}(x, t), .\}$ and the volume $f_y(x, z, t) = \{., f_{yz}(x, t), .\}$, the planes $f_{zy}(x, t)$ and $f_{yz}(x, t)$ behave differently since they are optimized independently to capture time variations in the volumes $f_z(x, y, t)$ and $f_y(x, z, t)$, respectively. In this way, the 9 planes can leverage time-aware information in all possible combinations of the time and spatial dimensions ($xt, yt, zt, xyt, yzt, xzt, xyzt$). This property enables our representation to capture various dynamic information hierarchically at different levels of spatial dimensions. Therefore, our method is more efficient in representing dynamic scenes, where variations in time dimension are often intense, complex, and long-range.

Our decomposition supports various types of 4D fields including the time-conditioned radiance field in NeRF-T and the 4D flow field in D-NeRF. In this paper, we present a dynamic radiance field decomposition method (Sec.~\ref{sec:t4d_multiview}) for multiview dynamic reconstruction as well as a 4D flow decomposition method (Sec.~\ref{sec:t4d_mono}) for single-view setting.

\subsection{Coarse-to-fine Strategy}
\label{sec:coarse_to_fine}
To further improve the efficiency of our 4D decomposition, we propose an optional coarse-to-fine strategy to factorize the 4D fields into different scales in different training phases. In coarse level, we adopt low-resolution feature planes ($128\times 128$) to decompose the 4D fields, which improves the robustness of the training process and achieves fast convergence. After coarse level training, we additionally use high-resolution feature planes ($512\times 512$) for 4D decomposition to represent dynamic details and achieve high-quality rendering. Specifically, we factorize the 4D fields into different scales:
\begin{equation}
\label{equ:coarse_to_fine}
f(x, y, z, t) = \left\{ \pi_3(\Pi_3(f^{LR})), \pi_3(\Pi_3(f^{HR}))  \right\},
\end{equation}
where $f^{LR}$ and $f^{HR}$ are the coarse-level and fine-level components of $f(x, y, z, t)$, respectively. 
In the coarse level, the decomposed feature planes is low-resolution to represent coarse 3D structures and 4D dynamic changes. In the fine level, we adopt the high-resolution feature planes  in each element to decompose the 4D fields, which focuses more on recovering dynamic details.

\begin{figure*}
    \centering
    \includegraphics[width=1.0\linewidth]{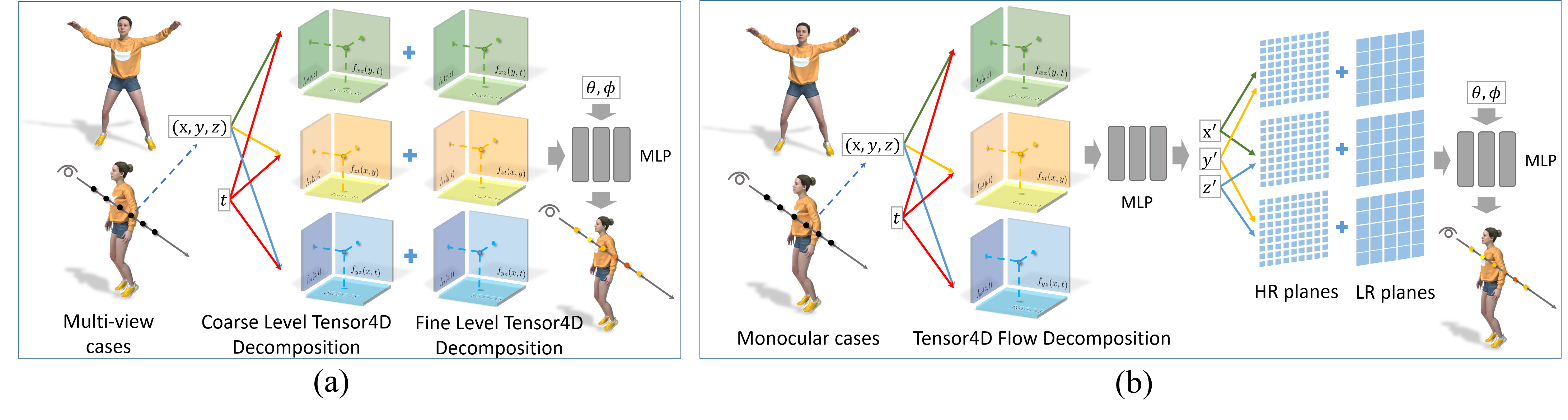}
    \caption{\textbf{The framework of Tensor4D for multi-view and monocular reconstruction.} \textbf{a).} Tensor4D for multi-view reconstruction. The 4D NeRF-T fields are separately factorized by the LR and HR feature planes. \textbf{b).} Tensor4D for monocular reconstruction. The 4D flow fields are factorized by the LR feature plane for better disentanglement of shape and motion. The 3D canonical representation is factorized by three LR and HR feature planes.}
    \label{fig:frame}
    \vspace{-5mm}
\end{figure*}

\section{Tensor4D for Dynamic Reconstruction}

As shown in Fig.~\ref{fig:frame}, we apply our 4D tensor decomposition into the task of dynamic reconstruction with two inputs: 

1) Dynamic reconstruction under sparse and fixed cameras. 
For this setting, 
we factorize NeRF-T instead of D-NeRF using our proposed 4D decomposition, which we found more efficient and flexible to represent topologically-varying objects (Sec.~\ref{sec:t4d_multiview}).  

2) Dynamic reconstruction using monocular camera. For this setting, we separately decompose the 4D flow fields and the 3D canonical representation since it ensures appearance consistency across different frames and achieves more robust performance (Sec.~\ref{sec:t4d_mono}).

\subsection{Multi-view Reconstruction}
\label{sec:t4d_multiview}
In this section, we present our dynamic reconstruction system under sparse multi-view setting. 
The goal of our system is 1) efficient and high-quality dynamic reconstruction with low memory and time cost, and 2) robust reconstruction under sparse and fixed camera setting. 

To this end, we adopt our 4D decomposition (Eq.~\ref{equ:coarse_to_fine}) to factorize the NeRF-T representation with the coarse-to-fine strategy. Specifically, we can obtain nine low-resolution feature planes $T^{LR}$ and nine high-resolution feature planes $T^{HR}$ to represent the 4D NeRF-T fields after our 4D decomposition. Then for volume rendering, when sampling a point $p$ at $(x, y, z, t)$ in the ray $r$ with direction of $(\theta, \phi)$, we first query the feature $F_p$ of $p$ in both the LR and the HR feature planes. Take the LR features for example:
\begin{equation}
\label{equ:query_feature}
\begin{split}
F^{LR}_p &(x, y, z, t) = \pi_3(\Pi_3(f^{LR})) \\
 &= \oplus(\pi_3(f^{LR}_z), \pi_3(f^{LR}_y), \pi_3(f^{LR}_x)) \\ &= \oplus(\oplus(T^{LR}_{zt}(x, y), T^{LR}_{zy}(x, t), T^{LR}_{zx}(y, t)), ...),
\end{split}
\end{equation}
where we  concatenate all features queried in the nine LR planes as the final LR feature of the sampling point $p$. 
Then we concatenate the LR and HR features with the positional encoding of $(x, y, z, t)$ into the geometry MLP $E_g$ to obtain the density $\sigma$ and high-dimension feature $\mathbf{f}$: 
\begin{equation}
    E_g(F^{LR}_p, F^{HR}_p, \gamma(x, y, z, t)) = (\mathbf{f}, \sigma),
\end{equation}
where $\gamma()$ is the positional encoding function. Next, we concatenate the high-dimension feature $\mathbf{f}$ with the positional encoding of $(\theta, \phi)$ and feed it into the color MLP:
\begin{equation}
\label{equ:color_predict}
    E_c(\mathbf{f}, \gamma(\theta, \phi)) = c
\end{equation}
In this way, we can render the images through volume rendering and adopt color loss to train our decomposed feature planes $T$ and neural network $E_c$ and $E_g$:
\begin{equation}
\label{equ:color_loss}
L_{c} =  \| C_{i, j} - C^{*}_{i, j} \|
\end{equation}
With our coarse-to-fine design, our method can recover high-fidelity dynamic details effectively and efficiently.

To achieve robust dynamic reconstruction under sparse multi-view setting, we further adopt regularization for all decomposed feature planes $T$:
\begin{equation}
\label{equ:feature_reg}
L_{r} = \sum_{T}\sum_{i, j}\sqrt{(T_{i+1, j} - T_{i, j})^2 + (T_{i, j+1} - T_{i, j})^2},
\end{equation}
where $L_r$ is the TV loss for each feature plane $T$ to keep their sparsity.
To regularize the geometry, we also introduce surface constraint into volume rendering. Specifically, we adopt the SDF as the base geometry representation and follow NeuS~\cite{neus2021} to render the SDF field. Then we add surface constraint loss to enforce a  smooth surface:
\begin{equation}
\label{equ:geometry}
L_{e} = \| \| \nabla s(x, y, z, t)\|_2 - 1 \|_2
\end{equation}
The final training loss $L_m$ is the regularization loss $L_{r},L_{e}$ and the color loss $L_c$:
\begin{equation}
\begin{split}
L_{m} & =   \lambda_rL_r + \lambda_eL_e + \lambda_cL_c     \\
\end{split}
\end{equation}

\begin{figure*}
    \centering
    \includegraphics[width=.95\linewidth]{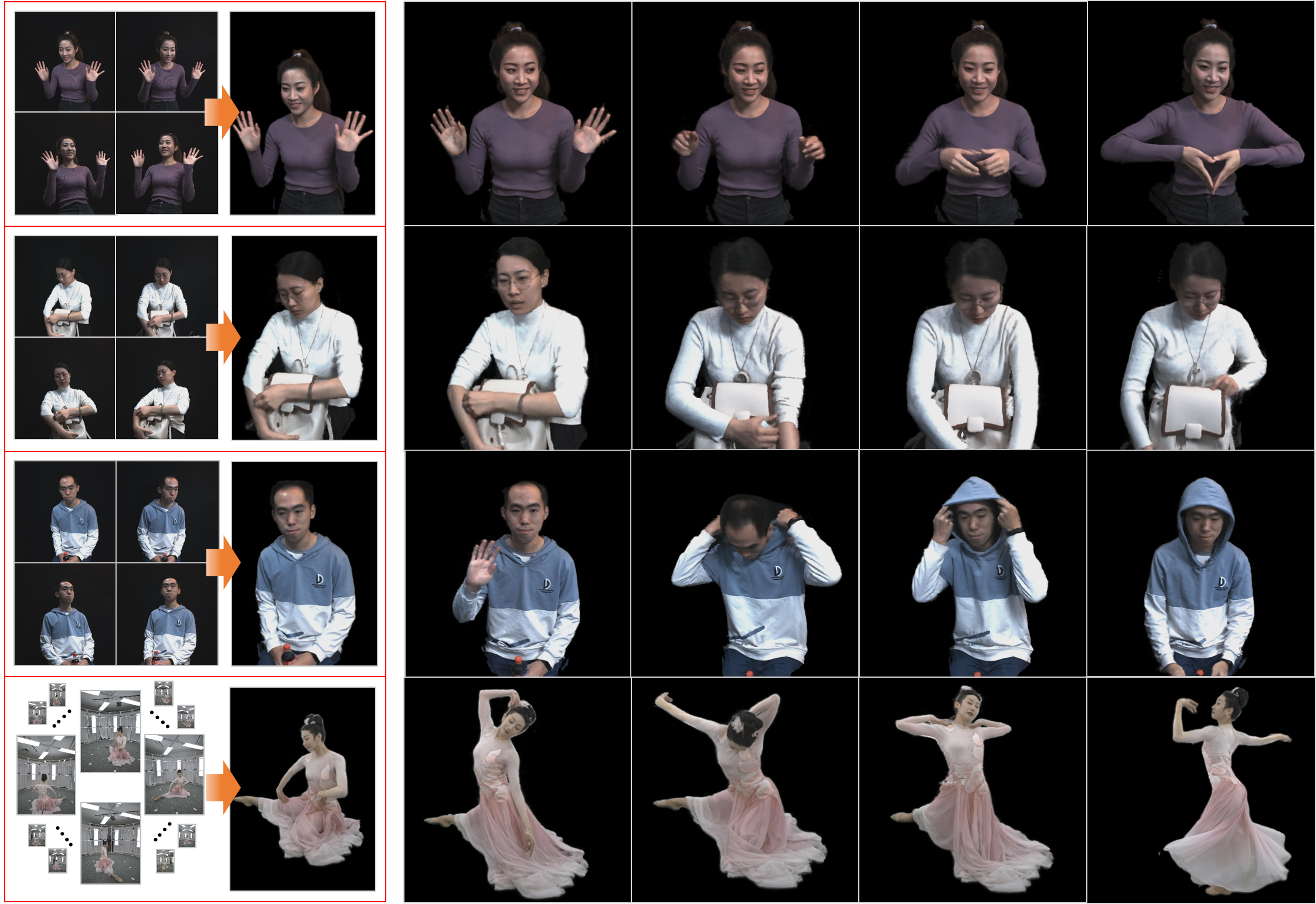}
    \caption{\textbf{Example results of our method.} Space and time novel view rendering results from sparse-view fixed cameras. The top three results are from four front view cameras and the bottom is from 12 circular cameras.}
    \vspace{-3mm}
    \label{fig:results}
\end{figure*}

\subsection{Monocular Reconstruction} 
\label{sec:t4d_mono}
Different from the sparse view setting, we adopt 4D decomposition for D-NeRF in monocular capture cases. This is because 
monocular setting is much more ill-posed than sparse-view inputs, and the explicit disentanglement of appearance and motions can guarantee the consistency across different frames. Since D-NeRF represents the dynamic objects with the 4D flow fields and a 3D canonical representation, we separately factorize these two fields. 
First, for the 4D flow fields, we only adopt coarse level decomposition and factorize it into low-resolution feature planes:
\begin{equation}
    f(x, y, z, t) = \pi_3(\Pi_3(f^{LR})).
\end{equation}
Our coarse decomposition focuses more on the coarse and rigid  motion, which improves the robustness of flow estimation and can achieve better disentanglement of shape and motion. Then for the 3D canonical representation, we adopt both coarse and fine level decomposition:
\begin{equation}
h(\hat{x}, \hat{y}, \hat{z}) \\ = \pi_3(h^{HR}) + \pi_3(h^{LR}).
\end{equation}
Therefore, we can obtain 9 flow feature planes $T_f$ for 4D flow fields  and 6 canonical feature planes $T_h$ for 3D canonical representation. The volume rendering in monocular cases is similar to multi-view cases. For a sampling point $p$, we first obtain the point flow feature $F_f$ using Eq.~\ref{equ:query_feature} with nine flow planes $T_f$. Then we adopt the flow MLP $E_f$ to predict the movement of the point:
\begin{equation}
E_f(F_p, \gamma(x, y, z, t)) = (\hat{x}, \hat{y}, \hat{z}).
\end{equation}
Then we obtain the point canonical feature $F_c$ by querying the canonical feature planes $T_h$. Take the LR feature $F_c^{LR}$ for example:
\begin{equation}
F_c^{LR} = \pi_3(h^LR) = \oplus(T^{LR}_{h_z}(x, y), T^{LR}_{h_y}(x, z), T^{LR}_{h_z}(x, y)).
\end{equation}
Next, we feed canonical feature $F_c^{LR}$ and the positional encoding of $(\hat{x}, \hat{y}, \hat{z})$ into the geometry MLP $E_g$ to obtain high-dimension feature $\mathbf{f}$ and density $\sigma$:
\begin{equation}
E_g(F_c^{LR}, \gamma(\hat{x}, \hat{y}, \hat{z})) = (\mathbf{f}, \sigma).
\end{equation}
Finally, we adopt Eq.~\ref{equ:color_predict} to predict color $c$ for volume rendering and Eq.~\ref{equ:color_loss} for training color loss $L_c$. We also add feature regularization loss $L_r$ in Eq.~\ref{equ:feature_reg} and surface constraint loss $L_e$ in Eq.~\ref{equ:geometry}. The total training loss $L_s$ is:
\begin{equation}
L_s = \lambda_cL_c + \lambda_rL_r + \lambda_eL_e.
\end{equation}

%% file: tex/experiment.tex
\begin{figure*}
\vspace{-3mm}
    \centering
    \includegraphics[width=0.97\linewidth]{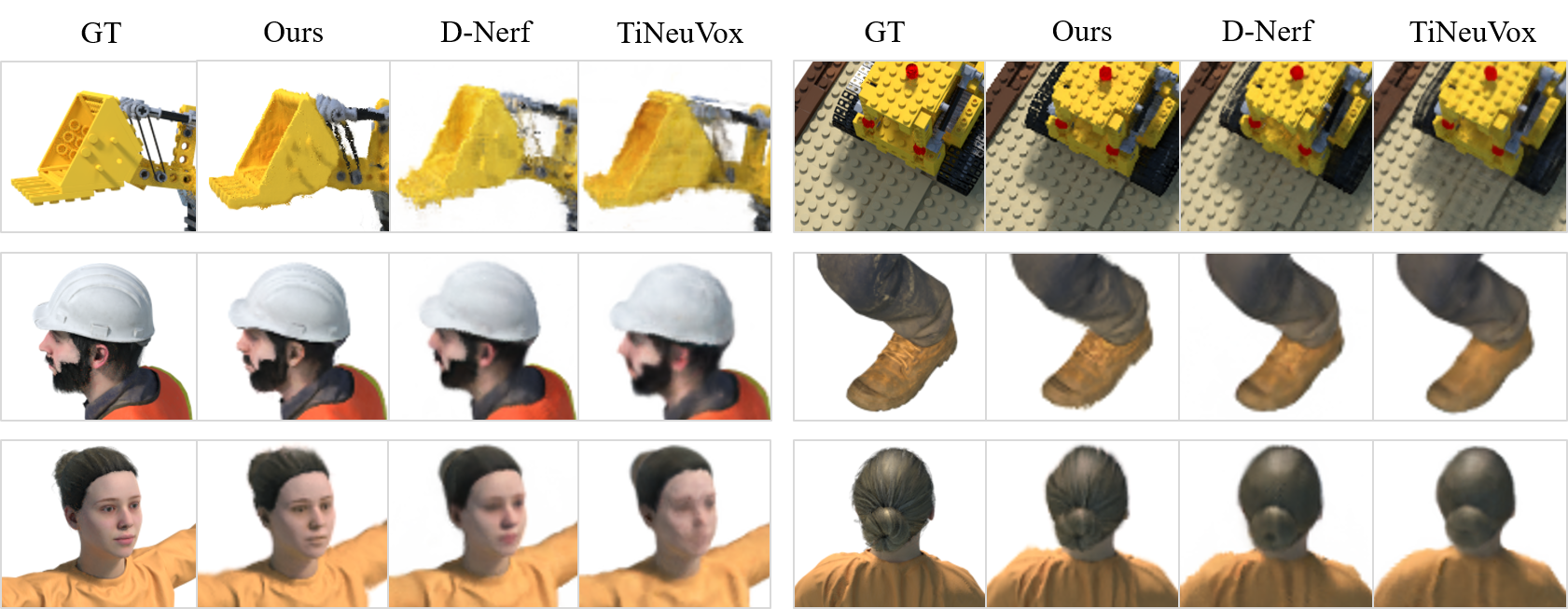}
    \caption{\textbf{Comparison on monocular synthetic dataset } against D-NeRF~\cite{D-Nerf2021-nfds54} and TiNeuVox~\cite{TiNeuVox2022-na13}. } 
    \label{fig:comparison_monocular}
\end{figure*}

\begin{figure*}
\vspace{-3mm}
    \centering
    \includegraphics[width=0.97\linewidth]{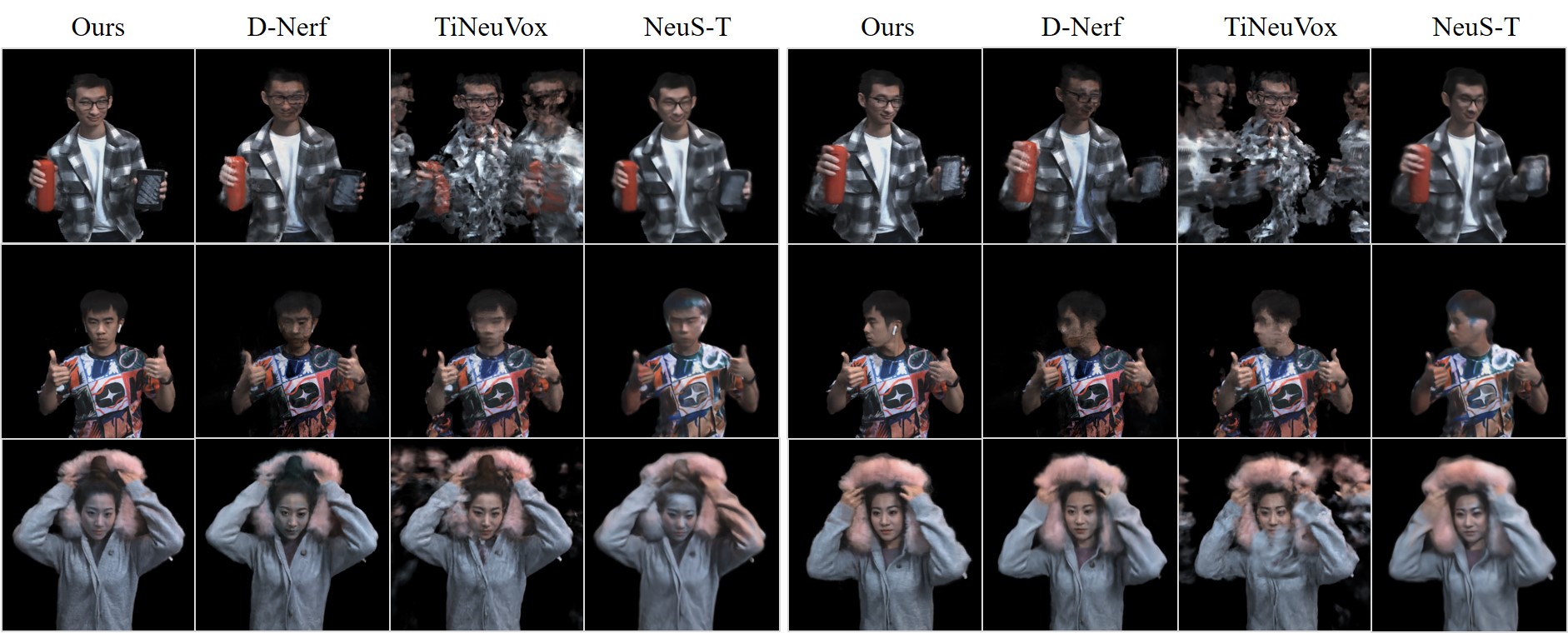}
    \caption{\textbf{Comparison on sparse-view real-world dataset } against D-Nerf~\cite{D-Nerf2021-nfds54}, TiNeuVox~\cite{TiNeuVox2022-na13} and Neus-T~\cite{neus2021}. Four of camera views from our datasets are used for reconstruction and rendering.}
    \label{fig:comparison_sparseview}
\end{figure*}

\begin{table*}[h]
   \footnotesize
   \centering
   
   \caption{Quantitative comparisons on monocular synthetic datasets. }
  \vspace{-1mm}
   \begin{tabular}{p{1.7cm}|cccc|cccc|cccc}
  \hline
  \multirow{2}*{\ \ \ Method} 
  & \multicolumn{4}{c|}{\makecell[c]{Lego}}
  & \multicolumn{4}{c|}{\makecell[c]{Standup}} 
  & \multicolumn{4}{c}{\makecell[c]{Jumpingjacks}} \\
         & MSE$\downarrow$ & PSNR$\uparrow$ & SSIM$\uparrow$ & LPIPS$\downarrow$ & MSE$\downarrow$ & PSNR$\uparrow$ & SSIM$\uparrow$ & LPIPS$\downarrow$ & MSE$\downarrow$ & PSNR$\uparrow$ & SSIM$\uparrow$ & LPIPS$\downarrow$ \\ \hline
        D-NeRF~\cite{D-Nerf2021-nfds54} & 7.83e-3 & 21.26 & 0.869 & 1.72e-1 & 4.63e-4 & 34.38 & \textbf{0.989} & 2.18e-2 & 5.46e-4 & 33.37 & \textbf{0.987} & 4.84e-2 \\
        NeRF-T  & 3.89e-3 & 24.32 & 0.904 & 1.55e-1 & 6.82e-4 & 31.44 & 0.968 & 2.36e-2 & 5.99e-4  & 32.27 & 0.979 & 5.37e-2 \\
       TiNeuVox~\cite{TiNeuVox2022-na13}   & 3.15e-3 & 25.14 & 0.924 & 8.37e-2 & 2.90e-4 & 36.18 & 0.986 & 2.02e-2 & \textbf{3.89e-4} & \textbf{34.76} & 0.983 & 3.33e-2 \\
      Ours-NeRF-T   & 3.10e-3 & 25.13 & 0.922 & 1.24e-1 & 5.92e-4 & 32.38 & 0.977 & 2.29e-2 & 5.62e-4 & 32.73 & 0.980 & 5.16e-2 \\
      Ours-D-NeRF   & 4.61e-3 & 23.37 & 0.890 & 1.19e-1 & 2.82e-4 & 35.93 & 0.981 & 2.07e-2 & 4.22e-4 & 34.10 & 0.982 & 3.41e-2 \\
    Ours  & \textbf{2.26e-3} & \textbf{26.71} & \textbf{0.953} & \textbf{3.49e-2}  & \textbf{2.50e-4} & \textbf{36.32} & 0.983  & \textbf{1.74e-2} & 3.91e-4 & 34.43 & 0.982 & \textbf{3.18e-2} \\
         \hline
  \end{tabular}
   \label{tab:comparison_monocular}
   
  \vspace{-3mm}
\end{table*}

\begin{table*}[h]
   \footnotesize
   \caption{Quantitative comparisons on the six camera-view real-world datasets.}
  \vspace{-1mm}
   \centering
   \begin{tabular}{p{1.7cm}|cccc|cccc|cccc}
  \hline
  \multirow{2}*{Method} 
  & \multicolumn{4}{c|}{\makecell[c]{Sequence1-thz}}
  & \multicolumn{4}{c|}{\makecell[c]{Earphone}} 
  & \multicolumn{4}{c}{\makecell[c]{Sequence3-yxd}} \\
         & MSE$\downarrow$ & PSNR$\uparrow$ & SSIM$\uparrow$ & LPIPS$\downarrow$ & MSE$\downarrow$ & PSNR$\uparrow$ & SSIM$\uparrow$ & LPIPS$\downarrow$ & MSE$\downarrow$ & PSNR$\uparrow$ & SSIM$\uparrow$ & LPIPS$\downarrow$ \\ \hline
        D-NeRF~\cite{D-Nerf2021-nfds54} & 3.13e-3 & 25.15 & 0.910 & 0.185 & 8.16e-3 & 20.92 & 0.854 & 0.256 & 4.84e-3 & 23.22 & 0.937 & 0.147 \\
       TiNeuVox~\cite{TiNeuVox2022-na13}   & 5.72e-3 & 22.79 & 0.832 & 0.209 & 1.49e-2 & 18.55 & 0.707 & 0.319 & 8.18e-3 & 21.19 & 0.816 & 0.233 \\
      Neus-T~\cite{neus2021}   & 4.95e-3 & 23.07 & 0.887 & 0.130 & 3.75e-3 & 24.22 & 0.877 & 0.184 & 2.20e-3 & 26.59 & 0.945 & 0.099 \\
      Ours-NeRF-T   & 5.85e-3 & 22.46 & 0.842 & 0.191 & 1.27e-2 & 19.13 & 0.838 & 0.218 & 5.02e-3 & 23.06 & 0.914 & 0.141 \\
      Ours-D-NeRF   & 4.74e-3 & 23.27 & 0.864 & 0.176 & 8.17e-3 & 21.07 & 0.883 & 0.198 & 4.06e-3 & 23.98 & 0.926 & 0.130 \\
    Ours  & \textbf{1.53e-3} & \textbf{28.27} & \textbf{0.942} & \textbf{0.084}  & \textbf{3.20e-3} & \textbf{25.00} & \textbf{0.903} & \textbf{0.153} & \textbf{1.31e-3} & \textbf{28.83} & \textbf{0.962} & \textbf{0.072} \\
         \hline
  \end{tabular}
   
  \vspace{-3mm}
   \label{tab:comparison_sparseview}
\end{table*}

\begin{table}[t!]
   \footnotesize
   \centering
   \caption{Training time and memory report in monocular cases. }
   \begin{tabular}{p{1.7cm}|cccc}
  \hline
  \multirow{2}*{Method} 
  & \multicolumn{4}{c}{\makecell[c]{Lego}}
  \\
         & PSNR & Time & Iterations & \#Params \\ \hline
         TiNeuVox   & 25.14 & 34min & 50k & 102M \\\hline
        D-NeRF & 21.26 & 45h & 800k & 4.8M \\
        Ours-D-NeRF  & 23.37 & 95min & 50k & 20M \\ \hline
        NeRF-T & 24.32 & 38h & 800k & 4.4M \\
      Ours-NeRF-T  & 25.13 & 73min & 50k & 43M \\ \hline
      Ours(6-planes)    & 26.34 & 135min & 50k & 17M \\
    Ours  & \textbf{26.71} & 144min & 50k & 20M \\
         \hline
  \end{tabular}
   \label{tab:efficiency_mono}
\vspace{-3mm}
\end{table}

\begin{table}[t!]
   \footnotesize
   \centering
   \caption{Training time and memory report in multi-view cases. }
   \begin{tabular}{p{1.7cm}|cccc}
  \hline
  \multirow{2}*{Method} 
  & \multicolumn{4}{c}{\makecell[c]{Sequence1-thz}}
  \\
         & PSNR & Time & Iterations & \#Params \\ \hline
         TiNeuVox   & 22.79 & 30min & 50k & 102M \\
        D-NeRF & 25.15 & 37h & 800k & 4.8M \\
        Neus-T    & 23.07 & 45h & 800k & 5.0M \\
      Ours-NeRF-T  & 22.46 & 68min & 50k & 43M \\
      Ours(6-planes)  & 27.86  & 105min & 50k  & 32M \\
    Ours  & \textbf{28.27} & 117min & 50K & 43M \\
         \hline
  \end{tabular}
   \label{tab:efficiency_multi}
\vspace{-3mm}
\end{table}

\begin{table}[t!]
   \footnotesize
   \centering
   \caption{Ablation study of hierarchical decomposition and smoothness terms.}
   \begin{tabular}{p{2.4cm}|cc|cc}
  \hline
  \multirow{2}*{Method} 
  & \multicolumn{2}{c|}{\makecell[c]{Lego}}
  & \multicolumn{2}{c}{\makecell[c]{Standup}} 
  \\
          & PSNR & SSIM  & PSNR & SSIM \\ \hline
      Ours(w/o regular)   & 26.49 & 0.946 & 35.91 & 0.977 \\
      Ours(6-planes)    & 26.44 & 0.944 & 35.79 & 0.978 \\
    Ours  & \textbf{26.71} & \textbf{0.953}  & \textbf{36.32} & \textbf{0.983} \\
  \hline
  \multirow{2}*{} 
  & \multicolumn{2}{c|}{\makecell[c]{Sequence-thz}}
  & \multicolumn{2}{c}{\makecell[c]{Sequence-earphone}} 
  \\
        & PSNR & SSIM & PSNR & SSIM \\ \hline
      Ours(w/o regular)   & 27.92 & 0.932 & 24.75 & 0.885 \\
      Ours(6-planes)    & 27.74 & 0.934 & 24.39 & 0.889 \\
    Ours  & \textbf{28.27} & \textbf{0.942} & \textbf{25.00} & \textbf{0.903} \\
         \hline
  \end{tabular}
   \label{tab:decomp_multi}
\vspace{-5mm}
\end{table}

\section{Experiment}

\noindent \textbf{Dataset.}
To evaluate the performance of our methods for multiview inputs, we build a sparse-view capture system with 6 forward-facing RGB cameras mounted on the borders of a 32'' Looking Glass 3D holographic display~\cite{LookingGlass}. All cameras are synchronized and calibrated. Using this system, we capture multiple sequences of various challenging human motions, including dancing, thumbing up, waving hands, wearing hats and manipulating bags. We use 4 of them for reconstruction and rendering in all our experiments, while leaving the other two for quantitative evaluation. We also use three 360° multiview full body sequences captured with $12$ evenly spaced cameras on a camera ring for qualitative evaluation. 
For monocular evaluation, we use the synthetic dataset provided by D-NeRF~\cite{D-Nerf2021-nfds54} and select 3 scenes (``lego'', ``standup'' and ``jumpingjacks'') from this dataset, with the numbers of training frames ranging from 50 to 200. 
More details about data collection and preprocessing can be found in the \textit{Supp.Mat.}.

\noindent \textbf{Baselines.} We mainly compare our method against the following state-of-the-art baselines that are most related to our work: D-NeRF~\cite{D-Nerf2021-nfds54}, NeRF-T, TiNeuVox~\cite{TiNeuVox2022-na13} and NeuS-T. Here, NeRF-T is our extension of vanilla NeRF~\cite{nerf2020} by introducing an additional time input, and NeuS-T is extended from NeuS~\cite{neus2021} similarly.  Among these baselines, D-NeRF and TiNeuVox represent the dynamic scenes through deforming a canonical one, while NeRF-T and NeuS-T directly learn a time-conditioned 4D radiance fields. TiNeuVox uses explicit voxel grids to accelerate network training, while others purely use MLPs to model the scene. 

\subsection{Results and Comparison}
\noindent \textbf{Qualitative Results.} 
We train our model for each individual sequences, and present some example results for novel view synthesis in Fig.~\ref{fig:results} and the \textit{Supp.Mat.}.
The results cover various body motions, clothing styles and accessories. As shown in Fig.~\ref{fig:results}, 
our method can render high-quality images for dynamic scenes and faithfully recover appearance details like the thin finger motions, semi-transparent silk, hand-object interaction, face expressions and cloth wrinkles. See our \textit{Supp.Video.} for better visualization.

\noindent\textbf{Comparisons on monocular dynamic dataset.}
We first compare our method with the baselines on monocular synthetic dataset. Qualitative results are presented in Fig.~\ref{fig:comparison_monocular}. Compared to other methods, ours recovers more appearance details and generate less artifacts. The numeric results in Tab.~\ref{tab:comparison_monocular} also prove that our method outperforms state-of-the-art methods in terms of rendering quality and accuracy.

\noindent\textbf{Comparisons on sparse view dataset.}
We then evaluate the performance of different methods for novel view synthesis given four camera views from our collected six camera-view real-world datasets. The remained two views are used for quantitative evaluation. Results reported in Fig.~\ref{fig:comparison_sparseview} and Tab.~\ref{tab:comparison_sparseview} show again that our method performs better in accurate and high-quality appearance detail synthesis.

\noindent \textbf{Comparisons of training efficiency.}
We compare the training time and model size for memory in Tabs.~\ref{tab:efficiency_mono} and~\ref{tab:efficiency_multi}. 
Our method requires significantly less training time compared to the original D-NeRF, NeRF-T, and Neus-T in monocular and multi-view scenarios.
Compared to Ours-NeRF-T and Ours-D-NeRF, the longer training time observed in our method is due to the additional computation for geometry smoothness terms, which reduces artifacts and enhances rendering quality.
Compared to TiNeuVox, our method has a reduction in memory consumption and a clear improvement in rendering quality since Tensor4D enables the efficient decomposition of 4D fields at higher spatial resolutions ($512^3$ vs. $256^3$).

\subsection{Ablation study}

\noindent\textbf{Regularization.} 
We quantitatively ablate the regularization terms in our method. We implement two strong baselines ``Ours-NeRF-T" and ``Ours-D-NeRF", in which we directly apply our Tensor4D decomposition for NeRF-T and D-NeRF without the regularization terms. The results are reported in Tab.~\ref{tab:comparison_monocular} and Tab.~\ref{tab:comparison_sparseview}. Benefiting from our 4D decomposition, they achieve better performance than the original NeRF-T and D-NeRF. However, their rendering quality is still worse than our full method with regularization. In addition, We ablate for TV regular smoothness terms (Ours \vs Ours(w/o regular)) and the results are reported in Tab.~\ref{tab:decomp_multi}, which further validates the effectiveness of our smoothness terms for rendering quality enhancement.

\noindent\textbf{Hierachical Decomposition} We quantitatively ablate our hierarchical decomposition with 6-plane decomposition. As shown in Tab~\ref{tab:decomp_multi}, our method achieves superior performance in both multi-view and monocular cases, which validates the effectiveness of our hierarchical decomposition.

%% file: tex/discussion.tex
\section{Discussion}
\noindent\textbf{Limitations.} Since our method needs to decompose the 4D fields to several 2D feature planes, a pre-set bounding box of the scene is necessary. Therefore, it is difficult for our method to reconstruct backgrounds or objects which are out of the bounding box. Another limitation is our strong regularization terms. Though these terms benefit the robustness of our reconstruction under sparse views, they also limit our ability to handle challenging cases such as fluid and fog.

\noindent\textbf{Conclusion.}
We presented Tensor4D, a new method for learning high-quality neural representation for dynamic scenes from sparse-view videos or even a monocular video. To capture the spatio-temporal information in a compact and memory-efficient manner, we propose propose a novel hierarchical tri-projection decomposition method that models a 4D tensor with nine 2D feature planes. With proper design of training losses and regularization, our method provides an efficient yet effective solution to model the radiance fields of dynamic scenes. We believe our work can inspire future research towards low-cost, portable and immersive telepresence systems.

\paragraph{\bf Acknowledgements.}
This paper is supported by National Key R\&D Program of China (2022YFF0902200), the NSFC project No.62125107 and No.61827805 and the Beijing Science and Technology Project (Z211100004021006).